# Comparative study of 3D object detection frameworks based on LiDAR data and sensor fusion techniques

**Sreenivasa Hikkal Venugopala**[1,*]

[1]Master of Autonomous Systems, University of Applied Sciences Bonn Rhein Sieg, Sankt Augustin, Germany

hvsreenivasa93@gmail.com

**Abstract.** Estimating and understanding the surroundings of the vehicle precisely forms the basic and crucial step for the autonomous vehicle. The perception system plays a significant role in providing an accurate interpretation of a vehicle's environment in real-time. Generally, the perception system involves various subsystems such as localization, obstacle (static and dynamic) detection, and avoidance, mapping systems, and others. For perceiving the environment, these vehicles will be equipped with various exteroceptive (both passive and active) sensors in particular cameras, Radars, LiDARs, and others. These systems are equipped with deep learning techniques that transform the huge amount of data from the sensors into semantic information on which the object detection and localization tasks are performed. For numerous driving tasks, to provide accurate results, the location and depth information of a particular object is necessary. 3D object detection methods, by utilizing the additional pose data from the sensors such as LiDARs, stereo cameras, provides information on the size and location of the object. Based on recent research, 3D object detection frameworks performing object detection and localization on LiDAR data and sensor fusion techniques show significant improvement in their performance. In this work, a comparative study of the effect of using LiDAR data for object detection frameworks and the performance improvement seen by using sensor fusion techniques are performed. Along with discussing various state-of-the-art methods in both the cases, performing experimental analysis, and providing future research directions.

## 1. Introduction

Perception is the process of understanding the scene or environment by interpreting and organizing the information acquired through the sensory system. It is one of the basic building blocks of an autonomous system. In the case of autonomous vehicles, perception refers to the representation of the vehicle's environment. Perception system is categorized into various subsystems such as object detection and tracking system that is responsible for keeping a track of detected objects, mapping system responsible for building maps (both global and local) for path planning and obstacle avoidance, traffic signal detection system responsible for reading and communicating the signal information, and so on. A major challenge in scene understanding is the detection and tracking of the objects concerning the ego vehicle.

The perception system must be accurate, robust, and real-time [1], and to achieve this, autonomous vehicles are equipped with various automotive-grade sensors such as LiDAR, Radar, cameras, and ultrasonic sensors. Each sensor exhibits its superior qualities and inferior qualities under various circumstances. To exploit the complementary properties of these sensors, based on the different driving scenarios the autonomous vehicle chooses to fuse the multi-modal sensor information or to rely on a single sensing modality. Alongside, these sensors generate a huge amount of data that is difficult to be handled in real-time conditions. To this extent, deep learning techniques are employed for processing the incoming data and discarding the unnecessary data.

Various factors affect the performance of the perception system out of which a few of them pose major challenges. One such challenge is related to sensor limitations and adverse weather conditions. Sensors are the main source for acquiring information on the vehicle's environment, limitations in sensing the environment result in adverse effects on driving tasks which may lead to road accidents. An additional challenge is posed by the generalization capability of the system across various driving scenarios. A further possible technologically challenging scenario is related to obstacle avoidance. Another well-posed challenge is occlusion. Occlusion occurs when a certain object blocks another object which is in the view of the ego vehicle resulting in complete or partial visibility of the object. Along with this, the variations in the sizes of the object affects the sensor readings [2].

Leveraging the complementary information generated by the multiple sensors, deep learning-based 3D object detection methods generate a high-resolution representation of the environment by fusing the data from multiple sources and estimates the 3D location, size regressions, and orientations of the object. This helps in handling adverse weather effects, lighting conditions, decreasing the processing time of data, tackling occlusions, and avoiding collisions in real-time.

Recent advancement shows that fusing the data from multiple sources at different stages of processing results in rich representations which in turn supports accurate and precise 3D object detection. On the contrary, the LiDAR sensor can provide the 3D data and is not vulnerable to change in lighting conditions, and recent research such as SECOND [3], PointPillars [4], shows that the 3D detections performed using only LiDAR data involving deep learning techniques provide promising performance similar to sensor fusion-based methods.

In this regard, this research work presents a comparative study of 3D object detection frameworks for autonomous vehicles based on LiDAR data and sensor fusion techniques.

## 2. Sensors and sensor fusion

*2.1. Sensors*

Sensors are the major and primary source of information for the interpretation of the environment for an autonomous vehicle. These vehicles are equipped with various automotive graded sensors which include both exteroceptive and proprioceptive sensors. Due to the availability of a vast number of automotive graded sensors to choose from, various criteria such as range (short, mid, long ranges), high spatial resolution, robustness to climatic conditions, object classification capability, ability to measure speed, cost, size, and computational requirements.

In this chapter, we provide more information on various sensors used for perception in autonomous vehicles.

*1) Camera:* Images captured from cameras provide comprehensive information in their field of view. Different cameras namely monocular camera, stereo camera, thermal camera, time-of-flight (TOF) camera are used in autonomous vehicles. The information from these images is in the form of pixel intensities which provide the texture and shape of the object. This texture and shape information is necessary for classifying the road signs, various classes of objects, and lane markings.

A disadvantage of using a monocular camera is that it lacks in providing the depth information which is necessary for estimation of the size of the object and its position accurately. But this can be overcome by using the stereo camera setup. The depth information is calculated by applying matching algorithms to determine the concurrences in the images from both cameras. But this setup requires more processing time and energy. To overcome the consumption of processing time, the TOF cameras are used, but the downside of this camera is that the resolution is low when compared to the stereo cameras [5]. The weakness of cameras is that they are sensitive to lighting conditions and weather conditions.

*2) LiDAR:* LiDAR sensors generate high-resolution data by emitting and receiving thousands of laser beams. These laser beams hit the objects and bounce back to the receiver in the sensor. The time of flight of these laser beams is used to determine the distance of the object from the source. Based on the intensity of the reflected beams, the texture of the objects can be obtained. Each of these laser beams is in the infrared spectral range and are emitted in many different angles covering 360° generating highly accurate models of the vehicle's surroundings in 3D.

As these are active sensors, these do not require any external illumination factors, so these sensors are more reliable during nighttime as well as in case of sunny weather conditions. The main problem with LiDAR is that they generate a very large amount of data in less time.

One tricky behavior of the LiDAR sensor is that, if two sensors with the same wavelength are used, they tend to generate fake echoes of the objects resulting in the detection of objects nearer or further than the actual position. As the range of the laser beams increases, the sparsity of the data increases and does not provide the fine texture of the object.

*3) Radar:* Radar works on the concept of the Doppler effect, by measuring the runtime of the radio signal, the radial velocity of the object is determined. The distance between the object and the ego vehicle can also be measured by utilizing the total time taken by the radio wave to hit the object and reflect to the receiver. Due to the high penetrable nature of radio waves, they offer good performance under varying weather conditions. Since Radar uses shorter wavelengths, the smaller objects will not be detected and due to low resolution, the fine texture of the objects will not be generated, and it is very challenging to use Radar for the classification of the objects.

*4) Ultrasonic:* These use high-frequency sound waves in order to determine the object's distance. These are mainly used in the detection of objects that are near the ego vehicle and also in scenarios of low-speed driving viz automated parking. These sensors are very sensitive and the accuracy can be easily affected by a change in temperature, dust in the air, and also these waves are absorbed by soft materials such as fabrics.

*2.2. Sensor fusion techniques*

Sensor fusion is the integration of data from multiple sources in order to provide or generate a specific and comprehensive unified representation of the data [6]. Sensor data fusion helps in the improvisation of measurements from more than one source of data than individual sources of data. Sensor fusion helps in handling redundant data and increases system accuracy along with improving the integrity of the system [7]. Sensor fusion can be adapted in many ways depending on the time of fusion, and sensor data can be integrated on various levels, say, data level, feature level, and decision level.

Here data-level fusion refers to integrating the sensor data directly, feature level fusion refers to combining the representative features from multi-sensor data, and decision level fusion refers to fusing the decision from each sensor modal [8]. Sensor fusion is a multidisciplinary task and based on its application, data sources, and other criteria it can be classified into many categories, and detailed information on this is provided in the literature [9]. In this report, sensor fusion techniques for the application of the perception system are divided into three categories, (1) Probabilistic approaches, (2) Gaussian approaches, and (3) Deep learning approaches.

*1) Probabilistic approaches:* Probabilistic approaches generally model the uncertainties using probability theory. Here we briefly discuss the widely used Bayesian filters (Kalman filter [10]) for sensor fusion. Given the prior information related to the system and the measuring devices, along with all the measurement data from the sensors, Kalman filters provide an optimal approximation of the system state. Because of the filtering capability and modeling the uncertainties, Kalman filters are widely used in many applications especially in autonomous navigation systems for accurate localization of the system and tracking of the dynamically moving objects.

The Kalman filter consists of two models, one is the process model, and the other is the measurement model. The process models, by assuming the future state as a linear function of the present state, describes the transition of the process state and it is defined in the below equation.

$$x_t = Ax_{t-1} + Bu_{t-1} + w_{t-1} \qquad (1)$$

where, $x_t$ - system state at time t, A - state transition matrix, B - relates to optional control input $u_t$ to state, $w_t$ - process noise.

The measurement model provides the measurement at time step t, by assuming the observations as a linear function of the state and is defined in the below equation.

$$z_t = Hx_t + v_t \qquad (2)$$

where, $z_t$ - measurement at time t, $x_t$ - system state at time t, H - relates state to measurement, $v_t$ - measurement noise.

Here $w_t$ and $v_t$ are independent Gaussian white noises. Along with the estimation of the system states, Kalman filters also provide the posterior estimation of error covariance and are defined in the below equation.

$$P_t = (I - K_t H)P_t^- \qquad (3)$$

where, $P_t$ - posteriori estimate of error covariance, $P_t^-$ - priori estimate of error covariance, $K_t$ - gain matrix.

Unfortunately, the system states and their transitions are dynamic in nature and are rarely linear, to solve this nonlinearity we use the extensions of Kalman filters such as Extended Kalman filters (EKF) and Unscented Kalman filters (UKF). EKF and UKF both linearize the non-linear states and perform the state estimations. EKF uses Taylor series expansion techniques for linearizing the nonlinear system states and UKF performs stochastic linearization using a weighted statistical regression process. But the computational complexity of both these approaches is high.

*2) Gaussian approaches:* Gaussian processes are referred to as stochastic processes where any random subset of a random variable is jointly Gaussian distributed [11]. These are non-parametric Bayesian models and have continuous representation that helps in modeling the uncertain data and spatially correlated data. Gaussian processes are mainly characterized by a mean function and a covariance function or also known as a kernel that helps in modeling the distributions. Given the dataset, these kernel functions model how closely the data is related.

The main idea of the Gaussian process is to introduce the prior information directly in function space and it can be defined as a set of random variables with a finite number of data points that are Gaussian distributed. The introduction of this prior to the function space helps in representing the prior beliefs. Basically, combining this prior information along with the data results in posterior estimations.

Gaussian processes are also highly used in machine learning applications for classification and regression tasks along with dimensionality reduction tasks. Since these approaches are non-parametric models, they help in direct combination of functions or their derivatives. Initially, an approach for fusing multi-modal data was proposed by Pearlmutter et al [12], which performed a generalized transformation on the Gaussian process priors by utilizing the linear transformation, along with the use of transformed Gaussian priors to estimate the derivatives of the noisy measurements and fusing the data. A detailed report on various approaches, their performance along with advantages and disadvantages are provided in a literature review published by Vasudevan.

*3) Deep learning approaches:* For fusing the sensor data with deep learning approach, generally, we need to have firm answers for the following three questions, (1) What to fuse - tells about what data to fuse, (2) How to fuse - how to fuse the data and what operations to be implemented for fusing the data, and (3) When to fuse - this is similar to the three-stage of data fusion proposed by Hall et al. In this section, we will provide a brief summary of the various approaches by answering the above three questions.

**What to fuse:** Based on various approaches of processing the point cloud data and camera images, the following three approaches can be used for fusion of data, (1) fusion of feature maps extracted from 2D convolutions separately on point cloud data and RGB images, (2) fusion of feature maps obtained from point cloud data using 3D RPN and camera images, and (3) projecting camera images on to BEV representation of point cloud data or projecting LiDAR data on camera coordinates and perform object detection.

**How to fuse:** Fusion of data can be achieved by performing operations such as addition, average mean, concatenation, ensembles, mixture of experts.

**When to fuse:** Because of the hierarchical representation of features by deep neural networks, it offers many options for data fusion such as early fusion, late fusion, and middle fusion.

## 3. 3D object detection methods

The outbreak of research in the field of object detection using deep convolutional neural networks started when AlexNet [13] created a record by winning the ImageNet Large Scale Visual Recognition Challenge [14]. Generally, objects are detected and classified based on the intersection over union (IOU) of the ground truth data and the estimated data, and this estimation usually is based on the classification probability and bounding box regressions.

Since this research revolves around LiDAR data and sensor fusion techniques, information on image-based object detection approaches is not provided.

### 3.1. LiDAR based object detection

To overcome the lack of availability of direct depth measurements from cameras, an alternative sensor named LiDAR is used for 3D perception. Various methods have been proposed for solving the 3D object detection problems based on the 3D data from LiDAR sensors. This 3D data is often referred to as point cloud data. Based on the different approaches of processing this data and utilizing them for object detection, the 3D detection methods can be sub-categorized as projection-based, volumetric representations, and point-nets.

Projection-based detectors mainly work by projecting the 3D point cloud onto a 2D image plane. Projection onto a 2D image plane can be performed by projecting on three different 2D shapes, namely, a plane [15], a cylindrical surface [16], or a spherical surface [17]. After projecting onto these surfaces a state-of-the-art 2D detector is utilized for object detection and by performing the dimension and position regression the 3D bounding boxes are recovered. Some of the projection-based approaches are Complex-YOLO [18], BirdNet, FCN based approach by Bo Li et al, an approach for normalizing the density channel was proposed by Beltrn et al, probabilistic approach by Feng et al [19].

Due to the loss of information during the projection of a 3D point cloud onto a 2D image plane, the detection estimation of the methods based on projection is not reliable and also there will be no spatial information encoding explicitly.

The second way of representing the point cloud inputs is using volumetric representations, the methods which use this kind of representation generally assume that the object or the scene is described in a 3D grid, or using voxels, where each of the units will be associated with an attribute. Due to this representation, the methods will encode the information on the shape of the object explicitly. One of the downsides of this representation is that if objects are less in a scene, most of the volumes will be empty and the efficiency of the methods will be reduced while processing these empty volumes. A few of the volumetric representation methods are 3DFCN [20], Vote3Deep [21].

Generally, the total number of sparsely distributed 3D points in a point cloud will be fluctuating. The conventional pipelines of deep neural networks assume that the input will be of fixed size, to avoid this downside, and to reduce the loss of information in the point cloud by projecting techniques the third method called point-nets is used for the representation of point clouds. Some of the point-net based approaches are VoxelNet [22], SECOND, IPOD [23], PointRCNN [24], PointPillars[25].

Among all the ways of representing the point cloud, projection-based techniques are popular since they can be used with traditional image-based object detectors, but due to loss of information due to projection, the point-net-based methods that use raw point cloud are preferred.

### 3.2. Sensor fusion-based object detection

Discrimination of various classes of objects is necessary for the task of object detection and classification. The texture of the object is an important factor for class discrimination, and it is not provided by LiDAR and are available with camera images. On the contrary, a monocular camera does not provide depth information which is a major requirement for size and location estimation of objects. Similarly, the camera can aid in the detection of the objects that are far from the ego vehicle but as the distance increases the sparsity of the LiDAR points increases and reduces detection capabilities. To utilize and exploit these complementary properties of these sensing modalities many methods were proposed for fusing the data from different sensors and performing object detection.

Instead of fusing the data from camera and LiDAR directly, there are few approaches that perform a fusion of RGB images along with depth images and optical flow to perform object detection in 2D space. A few such methods were proposed by Gonzalez et al [26], Enzweiler et al [27].

Various approaches such as MV3D [28], AVOD [29], Deep Continuous Fusion [30], LaserNet++ [31], multi-task multi-sensor fusion [32], Frustum PointNet [33], Frustum ConvNet [34] performing 3D object detection based on multi-modal data fusion and providing promising results were proposed. Various sensor fusion techniques and different approaches mentioned earlier exploit the complementary information from various sensing modalities resulting in improving the accuracy of 3D object detection and bounding box regression tasks.

**4. Evaluation**

*4.1. Datasets*
The first large-scale dataset for the autonomous driving task which was released for public usage was the KITTI dataset [35]. For all the datasets that are available, the KITTI dataset is widely used, and it is the benchmark for most of the research. KITTI dataset was recorded using a station wagon by attaching various sensing modalities such as a high-resolution stereo camera (both color and grayscale), Velodyne 3D laser scanner (LiDAR), and high-precision GPS/IMU inertial navigation system [36]. The latest version of this dataset was published in the year 2015. In this dataset, objects are categorized into various classes, for instance, Car, Pedestrian, Person (sitting), Cyclist, Tram, and more. Here an almost equal amount of data for both training and testing purposes are provided. There are 7,481 training frames and 7,518 testing frames including 2D annotations, 3D annotations, and pixel-level segmentation information are available. The total size of the available dataset is 180GB.

The raw dataset is grouped into various categories for example road, city, residential, campus, and person. The complete dataset is divided into many categories depending on the sensor data and various task, for instance, there is a subset of data only for stereo evaluation, optical flow evaluation, scene flow evaluation, tasks such as depth completion and prediction, map building, object detection and tracking in 2D and 3D, road/lane detection, semantic/instance segmentation, and the raw data. The dataset was recorded during the daytime so there is not much variation in the lighting conditions of the data.

Based on the amount of visibility of the object and the height of the bounding box, the KITTI dataset provides three difficulty levels for measuring the performance of object detectors as given below:
1) Easy: Bounding box = Minimum height of 40px, Maximum occlusion level = Fully visible, and Maximum truncation level = 15%.
2) Moderate: Bounding box = Minimum height of 25px, Maximum occlusion level = Partly occluded, and Maximum truncation level = 30%.
3) Hard: Bounding box = Minimum height of 25px, Maximum occlusion level = Highly occluded, and Maximum truncation level = 50%.

*4.2. Metrics*
Evaluation of object detection techniques mainly relies on the calculation of precision and recall values. A trade-off between the precision and recall values of the various confidence values that are associated with the predicted bounding boxes by a model is observed through the precision v/s recall curve (PR curve). An object detector model is considered to be good if its precision stays high even when its recall values are high.

In practical applications, the PR curve will not be smooth and most of the time it will vary making it difficult to calculate the area under the curve. To overcome this, a metric called average precision was introduced. Average precision is a numeric metric that helps in comparing various detectors based on their values. The AP basically summarizes the shape of the PR curve and is calculated by taking the mean of precision values at a set of eleven equally spaced recall values as shown in the below equation [37].

$$AP = \frac{1}{R_{11}} \sum_{r \in \{0, 0.1, \ldots, 1\}} P_{interp}(r) \qquad (4)$$

where, $P_{interp}(r)$ - interpolated precision value, $R_{11}$ - 11-point interpolation points and is equal to 11.

Interpolation of precision at each recall level r is performed by considering the maximum precision value measured for which the corresponding recall value exceeds r as shown in the below equation.

$$P_{interp}(r) = \max_{r': r' \geq r} P(r') \qquad (5)$$

$P(r')$ is the measured value of precision at a recall value $r'$.

The main reason for 11-point interpolation is to minimize the effect of the zigzag effect of the PR curve. This is done to penalize the models that do not provide a precision value at all recall points. The main reason for introducing this metric is to compare different methods and to assign a rank based on the performance.

The average precision $AP_i$ over all the N classes in a dataset can be calculated by using the mean average precision (mAP) and is given as

$$mAP = \frac{1}{N} \sum_{i=1}^{N} AP_i \qquad (6)$$

Along with AP, average orientation similarity (AOS) is used to assess the performance of the 3D detection frameworks for detection and 3D orientation of the objects. AP bird eye view ($AP_{BEV}$) helps in evaluating whether the prediction of an object is accurate with respect to the ground-truth labels by performing the IOU of prediction with respect to the ground-truth from BEV space. This metric helps in evaluating the conditions for collision avoidance. $AP_{3D}$ counts the IOU of 2 cuboids one from prediction and other from the ground-truth and adds the information of the height of the object and up-down based on the object's location.

*4.3. Discussion*

A major drawback of the methods performing 3D estimation on monocular images is the lack of availability of depth information which limits the detection and localization of the objects especially when the objects are far or occluded and since monocular cameras are sensitive to changes in lighting and weather conditions the performance will be compromised. Due to these insufficiencies, the information on image-based object detectors is not discussed.

To perform the comparative analysis, PointPillars and F-ConvNet are chosen for conducting various experiments. PointPillars introduces a new approach for processing the raw point cloud by representing them in the column features called pillars and are two-dimensional vectors. These include a fast-encoding technique that encodes the raw point cloud into a pseudo image. This makes it easier and faster for performing 2D convolution operations and then an SSD-based detection head is utilized for detection and classification tasks. The inference time is too low and is faster than all previous approaches.

F-ConvNet fuses the frustum-level features extracted by grouping the point cloud based on the extraction of frustums that depends on 2D region proposals. Here, each 3D point is identified that corresponds to the pixel level information from 2D proposals, and frustums are generated. On these sequences of frustums, a series of PointNet is applied and the output is stacked into arrays on which using the FCN these frustum-level features are fused by performing convolution and sampling operations.

*4.4. Experiments and results*

For conducting the experiments, the scientific computing platform called cluster machines is utilized. These cluster nodes are equipped with 192 GB of RAM, Intel Xeon Gold 6130 (Skylake EP) processor with 16 cores, and Nvidia Tesla V100 GPU. For observing the impact on the performance of 3D object detection frameworks by using only LiDAR data over sensor fusion techniques various experiments were designed and conducted. Experiments such as general performance on the overall dataset, detection of occluded and truncated objects, the effect of change in range of LiDAR, and detection of multiple classes of objects using a single model.

Based on the quantitative and qualitative results published by PointPillars and F-ConvNet, the overall performance in detection of object class cars is quite similar for both frameworks and for detection of smaller objects like pedestrians and cyclists F-ConvNet produces better performance.

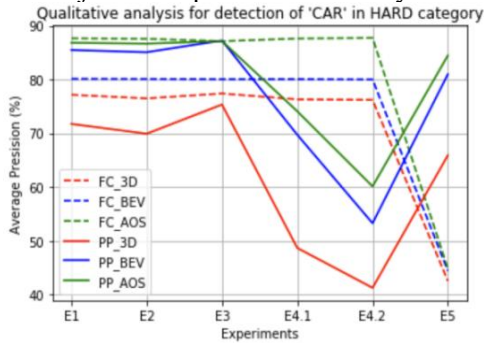

**Figure 1.** Performance graph for both frameworks across various experiments in detection of object class cars.

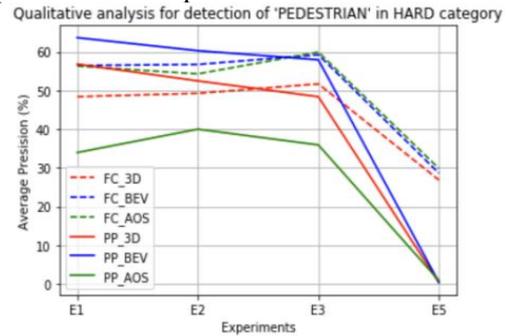

**Figure 2.** Performance graph for both frameworks across various experiments in detection of object class pedestrians.

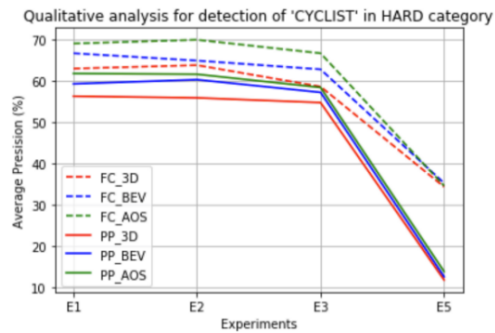

**Figure 3.** Performance graph for both frameworks across various experiments in detection of object class cyclists.

Along with the published results from both frameworks, in this work the performance in detection of object classes cars, pedestrians, and cyclists is compared by performing various experiments. As shown in Figure 1, 2, and 3, the x-axis represents different experiments (E1 - overall performance, E2- detection of occluded objects, E3 - detection of truncated objects, E4 - effect of change in LiDAR range on detection performance (E4.1 - 0-100m, E4.2 - 0-120m), and E5 - handling multi-class in a single model.) and the y-axis represents the average precision for the hard category.

The solid line in the graph represents the performance variation for PointPillars (PP_3D, PP_BEV, PP_AOS corresponds to the metrics $AP_{3D}$, $AP_{BEV}$, and $AP_{AOS}$) and the dashed line represents the performance variation for F-ConvNet (FF_3D, FF_BEV, FF_AOS corresponds to the metrics $AP_{3D}$, $AP_{BEV}$, and $AP_{AOS}$). We can see that the performance of F-ConvNet is not affected by variation in data and other conditions like occlusion, truncation, range of LiDAR, and the detection performance remains almost the same across different experiments, on the other hand, the performance of PointPillars degrades as the range increases, and also for the detection of smaller objects like pedestrians and cyclists.

The advantage of using LiDAR is that it provides depth information, and the scene can be visualized from a higher angle that majorly helps in avoiding occlusions and truncations. This together with features from the image supports better detection of objects. Based on the qualitative results shown in the above tables, it is seen that the performance of LiDAR-based framework on the detection of big objects such as cars is similar to that of sensor fusion framework even under various conditions and scenarios, but the performance with respect to detection of smaller objects is not quite comparable. The introduction of a new sensing modality such as a camera along with LiDAR improves the detection capabilities and the confidence level of the object detection framework.

**5. Conclusion**
In spite of the highlights and lowlights of the LiDAR sensor, the 3D object detection frameworks based on LiDAR data have proven to perform better in some scenarios, in the detection of bigger objects, and detection of smaller objects at a smaller range. But in real-time scenarios, it is not recommended to concentrate only on bigger objects or to keep varying the range of sensors for the detection of smaller objects. To overcome these lowlights of LiDAR frameworks, various sensor fusion techniques are discussed to show the improvement in object detection by utilizing the complementary effects of various sensors by fusing the data from multiple sources.

The textural information from RGB images helps in distinguishing between the pedestrian and a narrow vertical object even at a longer distance. This information is fused with the LiDAR data to improve the detection performance even at a longer distance. This fusion also helps in overcoming the sparsity issues associated with LiDAR. With fusion techniques, there is no need for varying the range of sensors for small object detection and also they provide real-time performance with high confidence values.

Additionally, there are a lot of areas for further improvement and future research perspectives and some of them are (1) Safety-critical points with respect to perception system in case of a sensor failure, noise in measurement, or effects of adverse weather conditions, (2) Implementing more sophisticated algorithms for the detection of smaller objects such as pedestrians in LiDAR data without confusing them with narrow objects.

**Acknowledgement**
I would like to acknowledge the support and guidance from Prof. Dr. Paul G. Plöger, Dr. Anastassia Kuestenmacher, Iman Awaad and Deebul Nair.